\documentclass{article}

\usepackage[english]{babel}

\usepackage[letterpaper,top=2cm,bottom=2cm,left=3cm,right=3cm,marginparwidth=1.75cm]{geometry}

\usepackage{amsmath}
\usepackage{graphicx}
\usepackage[colorlinks=true, allcolors=blue]{hyperref}
\usepackage{indentfirst}     
\usepackage{float}  
\usepackage{svg}
\usepackage{graphicx} 

\title{Integrated Pipeline for Monocular 3D Reconstruction and Finite Element Simulation in Industrial Applications}
\author{Bowen Zheng}

\begin{document}
\maketitle

\begin{abstract}
To address the challenges of 3D modeling and structural simulation in industrial environment, such as the difficulty of equipment deployment, and the difficulty of balancing accuracy and real-time performance, this paper proposes an integrated workflow, which integrates high-fidelity 3D reconstruction based on monocular video, finite element simulation analysis, and mixed reality visual display, aiming to build an interactive digital twin system for industrial inspection, equipment maintenance and other scenes. Firstly, the Neuralangelo~\cite{nerf} algorithm based on deep learning is used to reconstruct the 3D mesh model with rich details from the surround-shot video. Then, the QuadRemesh~\cite{rhino} tool of Rhino is used to optimize the initial triangular mesh and generate a structured mesh suitable for finite element analysis. The optimized mesh is further discretized by HyperMesh, and the material parameter setting and stress simulation are carried out in Abaqus to obtain high-precision stress and deformation results. Finally, combined with Unity and Vuforia engine~\cite{vuforia}, the real-time superposition and interactive operation of simulation results in the augmented reality environment are realized, which improves users 'intuitive understanding of structural response. Experiments show that the method has good simulation efficiency and visualization effect while maintaining high geometric accuracy. It provides a practical solution for digital modeling, mechanical analysis and interactive display in complex industrial scenes, and lays a foundation for the deep integration of digital twin and mixed reality technology in industrial applications.
\end{abstract}

\section{Introduction}

In industrial inspection and maintenance, accurate simulation and interaction with complex environments is essential to ensure safety, efficiency, and precision. However, traditional 3D modeling methods face significant limitations in hazardous or confined spaces due to the impracticality of deploying multisensor devices and the inherent trade-off between accuracy and real-time performance in existing techniques. For instance, while laser scanning offers high precision, it is time-consuming; in contrast, monocular vision enables real-time performance but produces lower quality results.To address these challenges, this project proposes a novel approach leveraging continuous monocular camera video streams to achieve high-fidelity 3D reconstruction. The goal is to overcome data acquisition limitations while balancing accuracy and real-time performance, ultimately generating digital models that support mechanical simulation and enable real-time interactive verification in mixed reality environments. This integrated pipeline enhances industrial capabilities by providing safer, more efficient, and precise solutions for simulating complex environments. 

Monocular 3D reconstruction is a core challenge in computer vision. Traditional approaches like Structure from Motion (SfM) rely on feature point matching and triangulation to achieve sparse scene reconstruction. However, SfM struggles in textureless or repetitive-texture areas, such as smooth walls or pipes. Moreover, it requires additional densification steps to generate usable models, increasing computational complexity. Recent advances in neural radiation fields (NeRF) have demonstrated remarkable capabilities in view synthesis, generating novel high-quality views from sparse images~\cite{sfm}. Nevertheless, NeRF's implicit representation (e.g., volumetric density fields) limits its direct application in physical simulations, as it cannot easily generate explicit meshes required for mechanical analysis. In the realm of simulation, traditional Finite Element Analysis (FEA) provides high-precision static simulations, but suffers from high computational costs, making it unsuitable for real-time applications. Conversely, game engine-based methods, such as Unity's physics engine, enable real-time interaction through simplified rigid body dynamics but lack the accuracy needed for complex deformations and material properties. While GPU-accelerated methods like NVIDIA PhysX~\cite{physx}. and deep learning-enhanced approaches (e.g., graph neural networks for deformation prediction) have improved the balance between accuracy and efficiency, a comprehensive solution integrating high-fidelity reconstruction with real-time, multi-scale simulation remains elusive. 

This project addresses the limitations of existing methods by developing a solution that integrates advanced monocular 3D mesh reconstruction with efficient multi-scale simulation for digital twin generation and interaction. Specifically, our contributions include: (1) a high-fidelity reconstruction pipeline leveraging Signed Distance Field for initial scene representation and view synthesis, followed by extracting explicit mesh models and optimizing irregular triangular meshes into structured quadrilateral meshes suitable for finite element analysis, enabling high-accuracy 3D models while overcoming the need of  multi-sensor devices; (2) high-precision finite element analysis (FEA) using Abaqus to simulate detailed mechanical behaviors such as material stress analysis and local deformation, with optimized quadrilateral meshes improving computational efficiency; and (3) mixed reality interaction utilizing the Vuforia engine to map the simulation results to mixed reality devices in real time, with Unity3D enabling AR display, scene marker recognition, visual overlay of simulation results, and user interaction support. By achieving these goals, our project provides a powerful toolset for industrial inspection and facility maintenance while advancing the practical application of digital twin technology in complex environments and offering new possibilities for mixed reality applications.

\section{Method}

\subsection{Architecture Overview}

\begin{figure}[H]
\centering
\includegraphics[width=\linewidth]{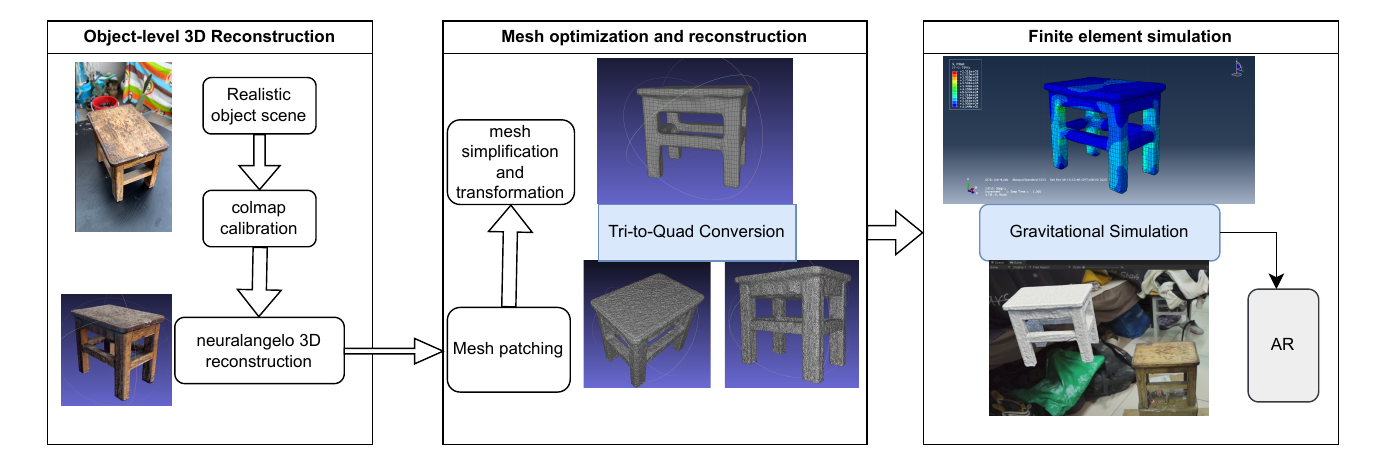}  
\caption{System architecture overview.}
\label{fig:system_architecture}  
\end{figure}

The architectural design of the project aims to obtain a high-quality 3D mesh model of an object through a multi-stage processing pipeline and convert it into a solid model suitable for FEA and mixed reality applications. The overall architecture consists of three main parts: 3D reconstruction and finite element analysis, and real-time visualization. Specifically,  the Neuralangelo is used to extract the fine mesh model of the object from the surround-shot video. The mesh is then repartitioned and optimized by Rhino~\cite{rhino} to generate high-quality quadrilateral meshes suitable for FEM analysis. Then, the optimized mesh was imported into HyperMesh~\cite{hypermesh} for further discretization, and the finite element mesh model was established and simulated by Abaqus~\cite{abaqus}. Finally, combined with the cloud map of the simulation results, it is visualized in real-time through the Unity platform~\cite{unity} to support more intuitive analysis and decision making. Through appropriate data exchange and integration technology each module can achieve seamless connection.

\subsection{3D reconstruction pipeline}

In this project, the core task of the 3D reconstruction pipeline is to recover a high-quality 3D mesh model of an object from surround-shot video. To achieve this goal, we employ the Neuralangelo algorithm~\cite{sfm}, which is based on deep learning techniques and is able to efficiently and accurately generate 3D geometry and texture information of objects from multiple viewpoints. The specific 3D reconstruction process includes the following key steps:

\begin{figure}[H]
\centering
\includegraphics[width=\linewidth]{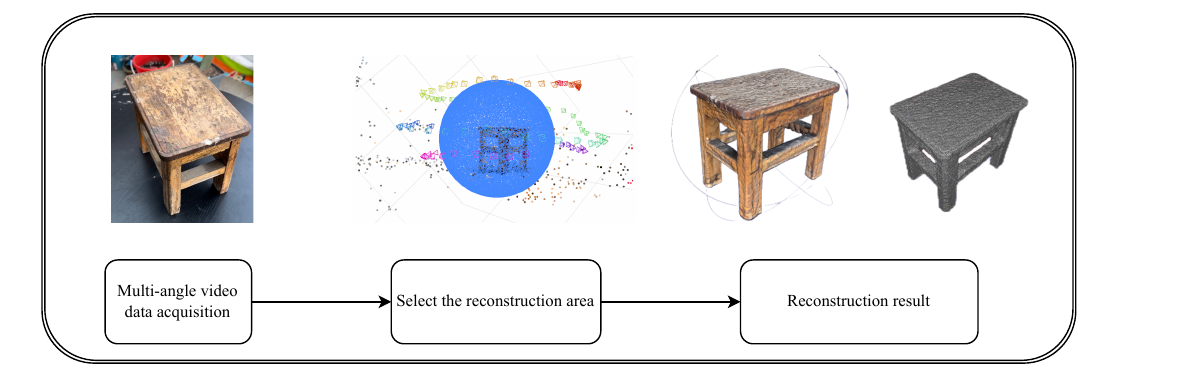}  
\caption{3D reconstruction.}
\label{fig:reconstruction}  
\end{figure}

Video Data Acquisition and Pre-processing: The first step of the project is to collect multi-view videos of the surrounding object. By shooting from different angles, multiple 2D images of the object are acquired, which contain various surface details of the object. Images data will be preprocessed before reconstruction,including image denoising, camera alignment, standardization, etc., to ensure the quality and consistency of the input data. We use colmap~\cite{sfm} for preprocessing to achieve camera extrinsic and intrinsic parameter calibration and extract sparse point cloud feature.
Feature extraction and Deep learning model training: On the basis of the preprocessed image data, the Neuralangelo algorithm determines the object location through the signed distance field. Deep learning models can automatically recognize object surface features,normals, and texture information in images. Unlike traditional 3D reconstruction methods, Neuralangelo does not rely on structured light or geometry matching from stereo vision. Instead, it learns the mapping relationship between images and geometry from different views through training a signed distance field function, and is able to preserve edge details Texture mapping and optimization: The Neuralangelo algorithm not only recovers the geometry of the object, but also extract the texture information to the 3D mesh accurately. By extracting texture features from multi-view 2D images, Neuralangelo ensures that the generated 3D meshes are both accurate in shape and highly fidelity in surface details. This process can truly reproduce the appearance of the object, including color, illumination and texture details, making the final mesh model more realistic.
High-resolution mesh output: After processing by the deep learning model, the final generated 3D mesh preserve high resolution and contains rich surface details. These meshes can not only accurately reflect the geometric characteristics of the object, but also can be used in subsequent applications such as mesh optimization and finite element analysis.

\subsection{Mesh optimization and FEA simulation}

\begin{figure}[H]
\centering
\includegraphics[width=0.7\linewidth]{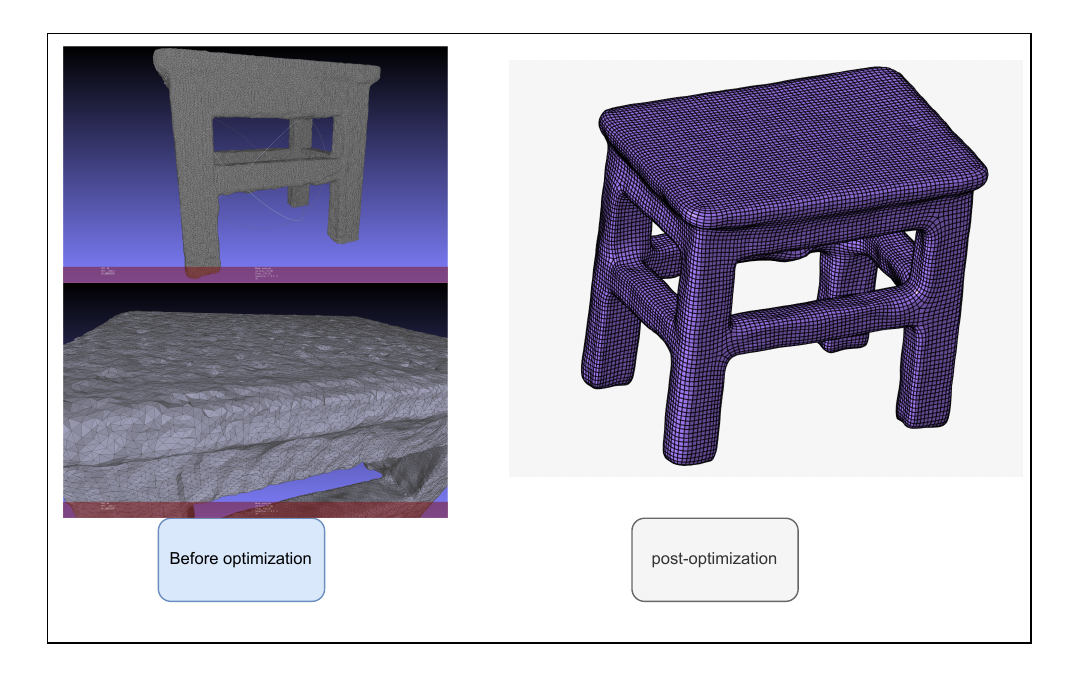}
\caption{Mesh optimization.}
\label{fig:optimization}
\end{figure}

To ensure suitable format for FEA analysis, we optimize the reconstructed object meshes and convert them in to quadrilateral mesh by using rhino Specifically,  we use QuadRemesh~\cite{rhino} command. The rationale behind the QuadRemesh command is based on a combination of local optimization and global mesh reconstruction, combining geometry analysis, Laplacian smoothing, and mesh generation algorithms. First, QuadRemesh analyzes each triangular element in the original mesh to determine the best mesh point location and node distribution when transforming the quadrilateral. This process makes the topology of the quadrilateral mesh more adapted to the curvature of the object surface. Moreover, it can effectively handle the special case of object boundaries, ensuring that the transformed mesh can correctly match the geometric boundaries of the object and avoid generating invalid or irregular meshes. The QuadRemesh algorithm adjusts the size and shape of mesh cells according to the curvature changes on the surface of the object. For regions with large curvature changes on the surface, the mesh will be finer. However, in regions with flat curvature, the mesh will be more sparse. This process can help maintain the geometric adaptability and quality of the mesh. The Laplacian smoothing algorithm is often used in the QuadRemesh generation process to reduce sharp corners, distorted or irregular mesh elements. Through several smoothing and subdivision operations, the mesh is gradually adjusted to make the final quadrilateral mesh more regular and uniform, and the distortion is reduced as much as possible.After that, the obtained mesh was converted into voxel representation using hypermesh~\cite{hypermesh}, which balance  accuracy and computational efficiency.
The voxel representation then is imported into Abaqus~\cite{abaqus} software and given the corresponding material properties and constraints to simulate the deformation and stress distribution of the object under different loads. The simulation results will provide a basis for the subsequent performance evaluation of the object.

\subsection{Mixed reality application}

\begin{figure}[H]
\centering
\includegraphics[width=0.8\linewidth]{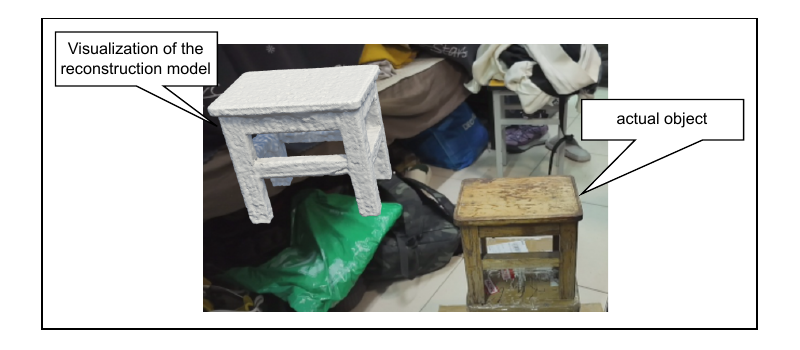}
\caption{Mixed reality application.}
\label{fig:Mixed_reality_application}
\end{figure}

After Abaqus simulation, the optimized high-quality mesh model is integrated into the augmented reality (AR) environments that enable visualization and interaction using Unity . First, object recognition is performed through the Vuforia engine~\cite{vuforia}, which uses computer vision techniques to detect and track predefined landmark, such as QR code in the real world, enabling the system to accurately localization and pose estimation. Then, the optimized 3D mesh is projected on the landmark within an augmented reality environment in real time to ensure an immersive visualization experience.

\section{Experiment}

\subsection{Evaluation}

To validate our envisioned architecture, we conducted reconstruction evaluation, simulation evaluation, and user experience evaluation, respectively.

\subsubsection{3D Reconstruction}

\begin{figure}[H]
\centering
\includegraphics[width=0.9\linewidth]{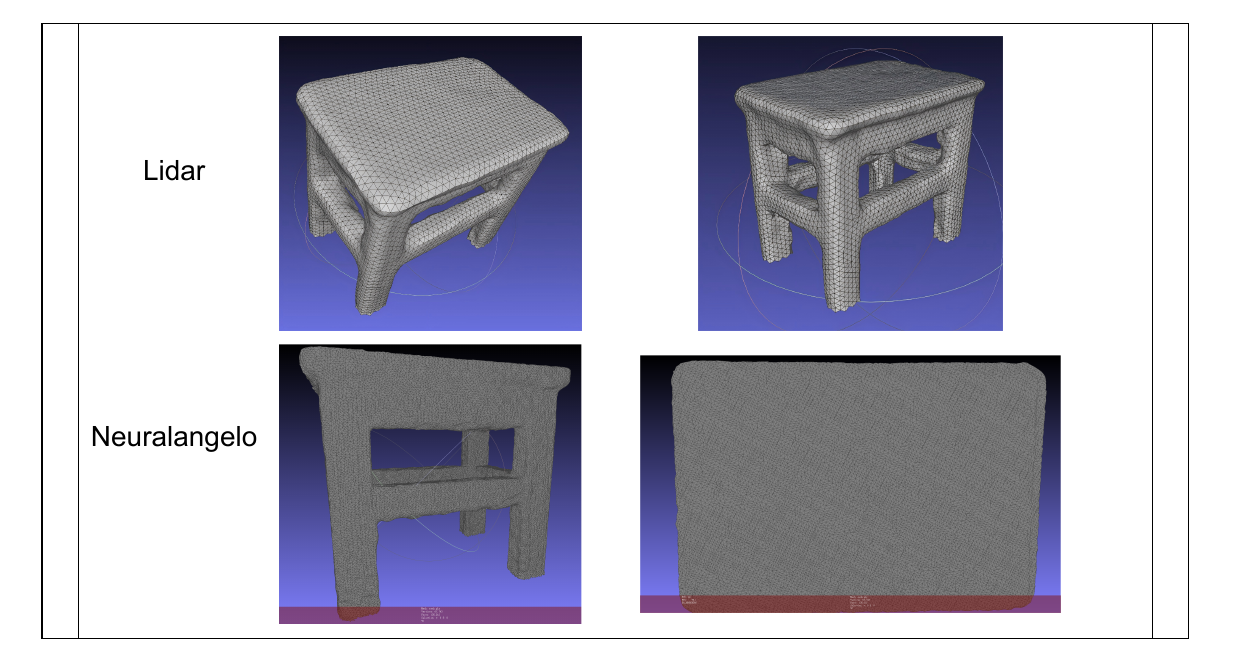} 
\caption{The meshes obtained from LiDAR and Neuralangelo.}
\label{fig:mesh_comparison} 
\end{figure}

In this study, we employed the built-in LiDAR sensor of the iPhone 14 Pro Max as a benchmark to conduct both qualitative and quantitative comparisons of 3D reconstruction performance. Although LiDAR systems are traditionally regarded as having high geometric accuracy, they still suffer from limited spatial resolution when applied to small-scale object reconstruction. This limitation manifests as a loss of fine details and blurred mesh boundaries. In contrast, our approach, which combines a monocular camera with a neural network-based method (Neuralangelo), achieves high-quality 3D reconstruction, particularly excelling in surface detail preservation.

For the qualitative analysis, visual comparisons of the reconstructed models indicate a high degree of structural similarity between the two methods. However, the LiDAR-generated model exhibits certain limitations in capturing fine geometric features, especially along object edges and small-scale textures, due to its inherent resolution constraints. Our method, on the other hand, visually preserves these geometric details more comprehensively.

To further assess geometric fidelity, Chamfer Distance was employed as a quantitative metric. The results demonstrate that our neural network-based reconstruction achieves comparable or even superior geometric consistency relative to the LiDAR-based model. These findings suggest that, despite the significantly lower hardware cost and simpler acquisition process of monocular camera systems, when combined with deep learning techniques, they can deliver reconstruction results that rival or exceed those of LiDAR systems, offering a compelling balance between performance and cost-effectiveness.

\paragraph{Chamfer Distance:}
To quantitatively assess the geometric similarity between the reconstructed 3D model and the reference LiDAR-scanned model, we employed the Chamfer Distance (CD) metric. Chamfer Distance is a widely used evaluation method in 3D vision tasks for comparing point clouds, as it effectively captures both global shape alignment and local surface fidelity.\\
Given two point clouds:
\[ P = \{p_i\}_{i=1}^N, \quad Q = \{q_j\}_{j=1}^M\]
representing sampled points from the reconstructed model and the LiDAR-based reference model respectively, the symmetric Chamfer Distance is defined as:
\[ \text{CD}(P, Q) = \frac{1}{|P|} \sum_{p \in P} \min_{q \in Q} \|p - q\|_2^2 + \frac{1}{|Q|} \sum_{q \in Q} \min_{p \in P} \|q - p\|_2^2\]
This formulation ensures bidirectional comparison, penalizing discrepancies in both directions between the two surfaces.In our experiment, in order to ensure statistical representativeness and consistency, we conducted uniform sampling on both 3D grids, and each grid had 100,000 surface points. Before calculating the distance, normalize the two point clouds to unit spheres to eliminate the influence of scale and shift differences.After conducting an efficient search by calculating the nearest neighbor distance using the k-d tree, the chamfer distance between the reconstructed model and the LiDAR model was obtained as 0.0561.This value represents moderate to high geometric similarity, indicating that the reconstructed mesh retains the overall shape and structure of the real objects captured by LiDAR. Although there are minor local differences (possibly due to surface noise, occlusion or untextured areas in the video input), the reconstruction shows satisfactory fidelity for downstream tasks such as simulation and visualization. In conclusion, the chamfer distance analysis verified the effectiveness of the proposed monocular reconstruction pipeline. Although relying solely on RGB video without depth sensing, this method has achieved an accuracy level comparable to that of active scanning technology, demonstrating its practical feasibility in cases where lidar is unavailable or impractical.

\subsubsection{FEA simulation}

In finite element analysis (FEA), mesh quality directly affects computational accuracy and stability. Therefore, a detailed mesh quality assessment was conducted before the simulation to ensure compliance with analysis requirements.\\
In this study, the mesh consists of both hexahedral and tetrahedral elements.\\
The specific statistics are as follows:

\begin{table}[H]
\centering
\caption{Mesh Quality Metrics}
\begin{tabular}{|l|c|c|}
\hline
\textbf{Metric} & \textbf{Hexahedral Mesh} & \textbf{Tetrahedral Mesh} \\
\hline
Total elements & 22,640 & 579,681 \\
Minimum angle \( < \) threshold & 634 (\( < \)10°, 2.8\%) & 0 (\( < \)5°) \\
Average minimum angle & 44.38° & 36.57° \\
Worst minimum angle & 4.68° & 7.68° \\
Maximum angle \( > \) 160° & 0 elements & 0 elements \\
Average maximum angle & 94.56° & 87.57° \\
Worst aspect ratio & 142.64 & 158.19 \\
Aspect ratio \( > \) 10 & 0 elements & 0 elements \\
Average aspect ratio & 1.43 & 1.77 \\
Worst aspect ratio & 3.62 & 8.21 \\
Shape factor \( < \) 0.0001 & -- & 0 elements \\
Average shape factor & -- & 0.6628 \\
Worst shape factor & -- & 0.0040 \\
\hline
\end{tabular}
\end{table}

\paragraph{Mesh Quality Analysis:}
Hexahedral Mesh: The overall quality of the hexahedral mesh is acceptable. The maximum angle does not exceed 160°, and the aspect ratio is well controlled. However, 2.80\% of the elements have a minimum angle below 10°, with the worst case being 4.68°, which may affect local computational accuracy. Therefore, mesh refinement in critical regions is recommended, such as adjusting seed size or employing an improved Sweep meshing technique.

Tetrahedral Mesh: For the tetrahedral mesh, all elements have a minimum angle greater than 5°, but the worst-case minimum angle is only 7.68°, and the worst shape factor is 0.004, which could lead to element distortion. Although no elements exceed an aspect ratio of 10, further mesh refinement is suggested in high-curvature regions to enhance computational stability.
\\
During the Abaqus Data Check process, there is no computational errors. Based on the mesh quality analysis, the current mesh is suitable for most static simulations.

\begin{figure}[H]
\centering
\includegraphics[width=\linewidth]{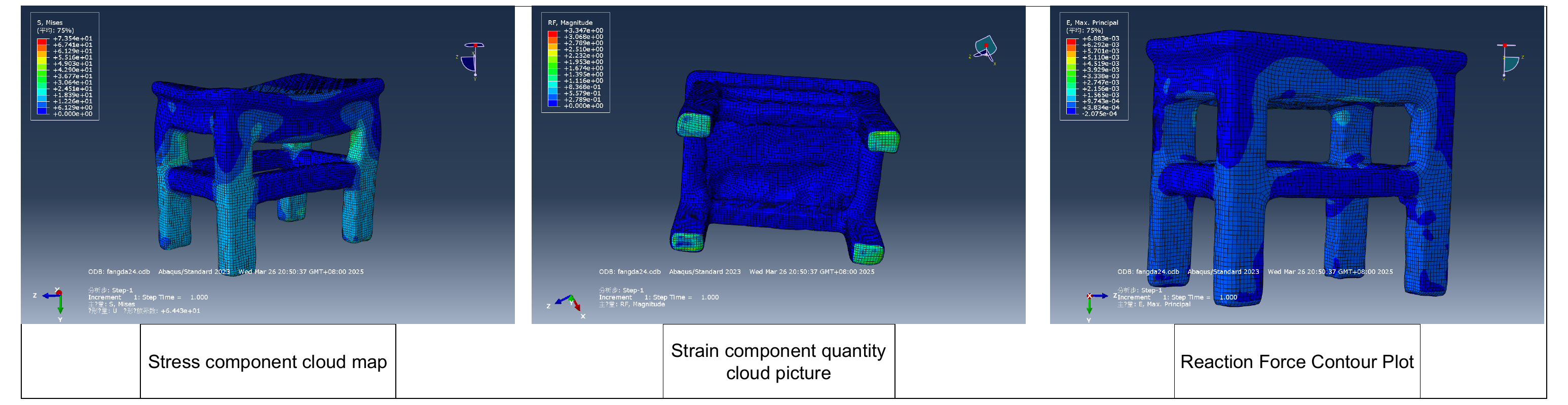} 
\caption{FEA Simulation.}
\label{fig:FEA_Simulation}
\end{figure}

\paragraph{Simulation Parameters:}

In this finite element analysis, the material used for the simulation is spruce wood, which has a density of $4.5 \times 10^{-7} \ \text{kg}/\text{mm}^3$, a Young’s modulus of 10,000 MPa, and a Poisson’s ratio of 0.3. These material properties were selected to accurately represent the mechanical behavior of the wooden structure under loading conditions.

For the loading conditions, gravity was first applied to simulate the realistic weight distribution of the stool in its natural state. Subsequently, a 500 N uniformly distributed load was imposed on the top surface of the stool to replicate the typical force exerted during usage.

Regarding boundary conditions, hinged constraints were applied at the bottom of all four legs to prevent translational movements while allowing rotational degrees of freedom. This setup ensures a realistic constraint condition, mimicking the interaction between the stool and the ground surface.

After ensuring the mesh quality meets the requirements and defining the simulation parameters, a finite element simulation was conducted to analyze the structural performance. The results include stress distribution, strain distribution, and reaction force contours, which provide insights into the mechanical behavior under the given load conditions.

Stress Distribution Analysis: The stress component cloud map presents the distribution of stress across the structure. The stress concentration regions are mainly observed at the joint connections and support areas, where the material undergoes higher loading. The maximum stress value is within the allowable limits of the material, indicating that the structure can withstand the applied forces without failure. However, the regions with higher stress values may require further design optimization, such as reinforcing critical areas or adjusting the load distribution to enhance durability.

Strain Distribution Analysis: The strain component cloud map shows the strain response of the structure under loading. The strain is distributed relatively evenly, except for certain localized regions where higher strain values appear. This suggests that the deformation mainly occurs in specific areas, which could be attributed to geometric features or material properties. The overall strain values remain within the elastic deformation range, ensuring the structure’s integrity under normal operating conditions.

Reaction Force Contour Analysis: The reaction force contour plot highlights the forces exerted at the constraints or support regions. The reaction forces are distributed symmetrically, confirming that the applied boundary conditions and loading setup are well-defined. The magnitude of reaction forces is consistent with theoretical expectations, further validating the correctness of the simulation setup. If excessive reaction forces are detected in certain regions, design modifications such as adjusting boundary constraints or redistributing loads may be necessary to reduce excessive localized forces.

\subsubsection{Mixed reality application} 
After completing the mesh quality analysis and simulation results analysis, in order to further improve the visualization and interactive experience of finite element simulation data, we use Unity to realize real-time recognition of modeling objects through AR overlay.

The workflow of the AR system consists of the following steps:

\begin{figure}[H]
\centering
\includegraphics[width=0.6\linewidth]{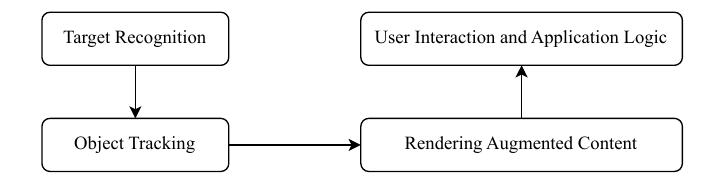}
\caption{Mixed Reality Application Workflow.}
\label{fig:Workflow}
\end{figure}

\begin{enumerate}
\item \textbf{Target Recognition:} \\
\hspace*{1em} The camera captures the real-world environment and continuously processes image frames. \\
\hspace*{1em} The frames are converted into a suitable pixel format for tracking. \\
\hspace*{1em} A predefined target database containing the object’s visual features is used to recognize the model.

\item \textbf{Object Tracking:} \\
\hspace*{1em} The Vuforia Tracker Module identifies and tracks the detected object in real-time. \\
\hspace*{1em} Features such as image targets, multi-image targets, and virtual buttons can be utilized for precise interaction and tracking.

\item \textbf{Rendering Augmented Content:} \\
\hspace*{1em} Once the object is detected, Unity renders the reconstructed mesh model in the AR environment. \\
\hspace*{1em} The model is displayed alongside the real object, allowing users to visualize the simulation results directly.

\item \textbf{User Interaction and Application Logic:} \\
\hspace*{1em} The application updates its logic based on the detected object’s state. \\
\hspace*{1em} Users can interact with the virtual model and analyze the finite element simulation results in an immersive manner.
\end{enumerate}

\begin{figure}[H]
\centering
\includegraphics[width=\linewidth]{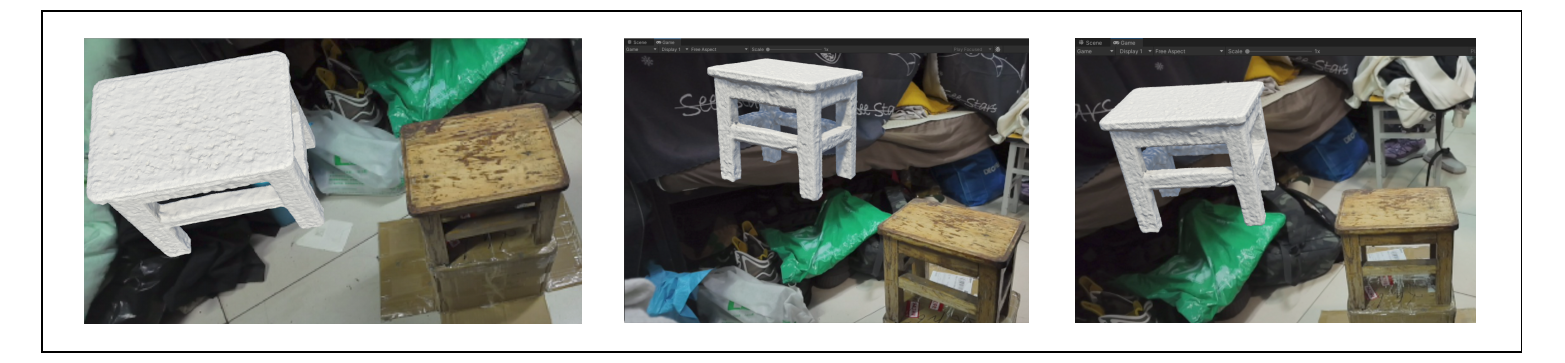} 
\caption{Virtual Reality Visualization.}
\label{fig:Visualization}
\end{figure}

The qualitative result is shown in Figure~\ref{fig:Visualization}.

\section{Discussion}
The proposed method offers a novel and effective solution to several long-standing challenges in industrial inspection and digital twin construction, especially in environments where traditional 3D modeling techniques are difficult to address. As mentioned in the introduction, existing methods usually face a trade-off between accuracy and real-time performance, especially in confined or dangerous Spaces where deploying multiple sensors is impractical. This study addresses these limitations by introducing an end-to-end framework that combines 3D reconstruction based on monocular video, mesh optimization, finite element simulation, and mixed reality visualization while maintaining a balance between high fidelity and efficiency.

Using Neuralangelo, a deep learning-based reconstruction algorithm, detailed surface modeling can be carried out from simple monocular input. Unlike traditional motion structures (SfM), sign distance field representation and neural rendering pipeline enable it to maintain geometric accuracy and retain textures under challenging conditions. In addition, the pipeline transitions from the original sparse point cloud to a clear high-quality quadrilateral mesh, facilitating the use of advanced finite element analysis (FEA) tools such as Abaqus. The typical problem of irregular triangular distribution in deep learning-based reconstruction was effectively solved through the intermediate mesh optimization step of Rhino's QuadRemesh, and it was transformed into a simulable structure. This simplified process can eventually enable accurate stress and strain analysis in a simulated mechanical environment.

Despite these advancements, there are still some limitations. During the reconstruction stage, areas with reflections, darkness or untextured surfaces still pose challenges to precise geometric restoration. These limitations can be attributed to suboptimal video quality or restricted viewing angles, which can introduce artifacts or lack geometry. Future iterations can solve this problem by combining hybrid data acquisition strategies, such as integrating depth information, such as the use of RGB-D camera to enhance robustness in difficult scenarios.

Another limitation lies in the current implementation of mixed reality (MR) interaction. The system relies on Vuforia’s image-based tracking, which is notably sensitive to variations in ambient lighting and subject to performance degradation under partial occlusion. These factors can compromise the accuracy of object tracking and alignment within real-world environments, reducing the reliability of the augmented overlay. Additionally, the current approach depends on predefined image markers for object recognition, which constrains the flexibility and scalability of deployment. Future improvements should focus on integrating markerless tracking techniques—such as feature-based spatial tracking—to enable more robust, flexible, and immersive MR experiences without the need for physical markers.

Furthermore, although finite element simulation has achieved a very high level of accuracy, its cost is a long computing time, especially when dealing with complex or high-resolution models. Although grid optimization partially alleviates this, further enhancements can be achieved by integrating machine learning-based agent models. For instance, graph neural networks (GNNS) trained to predict deformation patterns under various loads can significantly accelerate the simulation speed while maintaining acceptable accuracy, thereby supporting near real-time feedback in interactive scenarios.

\section{Conclusion}
This work proposes and verifies a comprehensive pipeline integrating monocular video 3D reconstruction, mesh optimization, high-precision finite element simulation and real-time mixed reality visualization. By leveraging deep learning techniques such as Neuralangelo for surface modeling and integrating the generated meshes with established FEA tools and AR platforms, this framework has successfully bridged the gap between data-driven reconstruction and physical simulation within a unified workflow. The result is a high-fidelity, simulable digital twin that supports immersive visualization and interaction, offering great potential for industrial inspection, maintenance, and virtual prototyping in complex environments.

\section{Implementation details}
In this study, we used NVIDIA's Neuralangelo for high-precision 3D mesh reconstruction. dict\_size = 21,dim=8. Rhino 8 \& QuadRemesh: used for grid optimization and repartition. HyperMesh 2023 was used for 3D mesh partitioning. Abaques CAE 2023 was used for finite element simulation. unity2022.3.47f1c1 and Vuforia Engine 11.1 were used for virtual reality display.

Neuralangelo is running on Google colab, using cuda 11.7 and python 3.10.16. The local computer graphics card version is 3070ti. Lidar is based on the Iphone 14 pro max.


\newpage
\begin{thebibliography}{8}

\bibitem{nerf}
Mildenhall, B., Srinivasan, P. P., Tancik, M., Barron, J. T., Ramamoorthi, R., \& Ng, R. (2020).  
NeRF: Representing scenes as neural radiance fields for view synthesis.  
In \textit{European Conference on Computer Vision (ECCV)} (pp. 405--421).  
DOI: \url{10.1007/978-3-030-58452-8_24}

\bibitem{rhino}
McNeel \& Associates. (2023). \textit{Rhino 8 Documentation: QuadRemesh Command}.  
Retrieved from \url{https://www.rhino3d.com/features/quadremesh/}

\bibitem{vuforia}
PTC. (2023). \textit{Vuforia Engine Developer Guide}.  
Retrieved from \url{https://developer.vuforia.com/}


\bibitem{sfm}
Schönberger, J. L., \& Frahm, J. M. (2016).  
Structure-from-Motion revisited.  
In \textit{Proceedings of the IEEE Conference on Computer Vision and Pattern Recognition (CVPR)} (pp. 4104--4113).  
DOI: \url{10.1109/CVPR.2016.445}

\bibitem{physx}
NVIDIA. (2023). \textit{PhysX SDK Documentation}.  
Retrieved from \url{https://developer.nvidia.com/physx-sdk}

\bibitem{hypermesh}
Altair Engineering. (2023). \textit{HyperMesh 2023 Documentation}.  
Retrieved from \url{https://altairhyperworks.com/hypermesh}

\bibitem{abaqus}
Dassault Systèmes. (2023). \textit{Abaqus CAE User’s Guide}.  
Retrieved from \url{https://www.3ds.com/products/simulia/abaqus}

\bibitem{unity}
Unity Technologies. (2023). \textit{Unity User Manual}.  
Retrieved from \url{https://docs.unity3d.com/Manual/index.html}


\end{thebibliography}
\end{document}